\documentclass{article}
\usepackage[preprint]{neurips_2026}
\usepackage[utf8]{inputenc}
\usepackage[T1]{fontenc}
\usepackage{graphicx}
\usepackage{hyperref}
\usepackage{url}
\usepackage{caption}
\usepackage{float}
\usepackage{wrapfig}
\usepackage{placeins}
\usepackage{needspace}
\usepackage{xcolor}
\usepackage{amsmath}
\usepackage{amssymb}
\usepackage{amsfonts}
\usepackage{booktabs}
\usepackage{multirow}
\usepackage{diagbox}
\usepackage{colortbl}
\usepackage{enumitem}
\usepackage{subcaption}
\usepackage{tabularx}
\usepackage{nicefrac}
\usepackage{microtype}

\usepackage{siunitx}

\title{\textcolor{black}{Predicting Multi-View Rashomon Representation: Can We Learn Where Models Disagree?}}
\author{
  \textbf{Mingyue Ma}\textsuperscript{1} \qquad
  \textbf{Zongbo Han}\textsuperscript{1}\thanks{Corresponding author: \texttt{hanzongbo.mail@gmail.com}.}\qquad
  \textbf{Changqing Zhang}\textsuperscript{2} \qquad
  \textbf{Guangyu Wang}\textsuperscript{1}\\[0.5em]
  \textsuperscript{1} Beijing University of Posts and Telecommunications \qquad
  \textsuperscript{2} Tianjin University
}

\begin{document}
\maketitle
\vspace{-1\baselineskip}
\begin{abstract}
\vspace{-0.5\baselineskip}
Foundation models are increasingly adopted across a wide range of applications, often serving as core blocks within AI systems. Yet different foundation models may encode the same input from multiple different views, leading to substantial representation disagreement, which we term \textbf{\emph{Rashomon Representation}}. Such disagreement often signals inputs that a given model encodes in a way inconsistent with other models, offering a valuable yet underexplored signal for input reliability estimation. While prior work has largely focused on \textbf{\emph{measuring disagreement}} across multiple models with a representation set, we instead focus on \textbf{\emph{predicting disagreement}} from a single representation. We hypothesize that this disagreement follows some consistent, input-dependent patterns rather than occurring at random. To test this, we quantify disagreement by comparing each sample's nearest neighbors across different models' representation spaces, then train a lightweight predictor that estimates disagreement from a single model's representation. At inference time, given a new input, the predictor uses that input's representation to tell whether it aligns with or diverges from those of other models. Extensive experiments across diverse foundation models and datasets show that representational disagreement is indeed input-dependent, predictable, and generalizable, enabling efficient reliability estimation of foundation models.\end{abstract}

\vspace{-1\baselineskip}

\section{Introduction}

\textcolor{black}{\emph{``The same event, told by different witnesses, becomes different events entirely.''}
\begin{flushright}
    --- \emph{Rashomon}, Akira Kurosawa, 1950
\end{flushright}}

\begin{wrapfigure}[18]{r}{0.44\textwidth}
    \centering
    \vspace{-0.8em}
    \includegraphics[width=\linewidth]{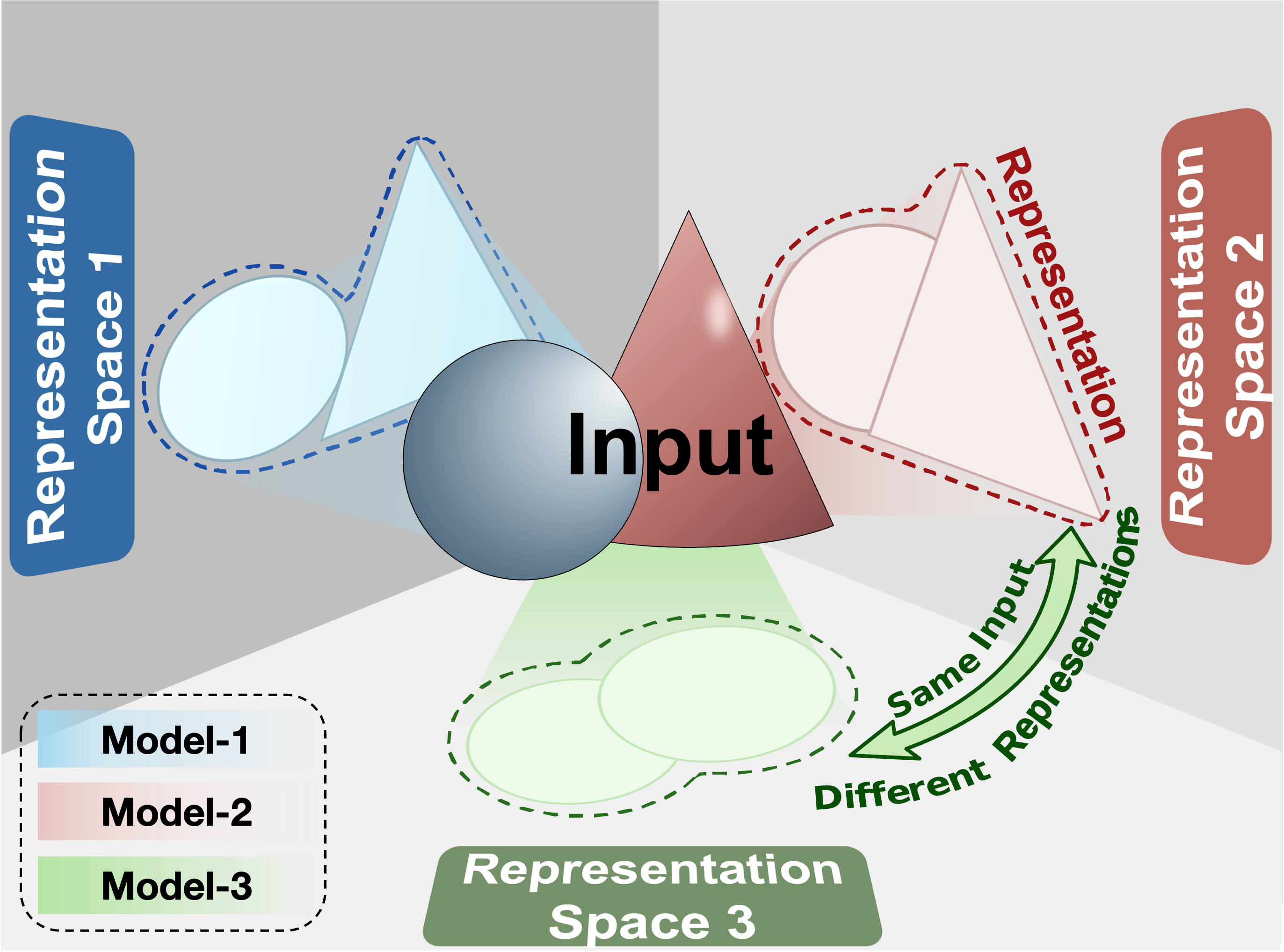}
    \caption{\textcolor{black}{Illustration of \emph{Rashomon Representation}: different foundation models can produce distinct representations of the same input, reflecting complementary views of its underlying information.}}
    \label{fig:overview}
    \vspace{-0.8em}
\end{wrapfigure}
\textcolor{black}{The rapid development of foundation models has enabled powerful and generalizable capabilities across a wide range of domains and applications \citep{awais2025foundation}. Models such as CLIP \citep{radford2021learning}, DINOv2 \citep{oquab2023dinov2} have demonstrated remarkable transferability, establishing a ``pretrain once, adapt everywhere'' paradigm. Despite their shared versatility, however, these models can learn fundamentally different representations. Differences in training objectives, architectures, and data distributions can lead them to encode the same input in substantially different ways \citep{pmlr-v267-ciernik25a,klabunde2025resi}. For example,  given an image, one model may focus more on high-level semantic attributes, while another may emphasize structural patterns \citep{barsellotti2025talking,zhang2025mamba}. As shown in Fig.~\ref{fig:overview}, different foundation models can produce distinct representations of the same input despite capturing similarly meaningful information. We refer to this phenomenon as the \emph{Rashomon Representation}. Such discrepancies reflect representation-level uncertainty that can affect downstream reliability \citep{park2023quantifying}. Understanding when they arise is therefore important for building robust and reliable AI systems.}

\textcolor{black}{Prior studies have largely focused on \textbf{measuring representation differences} through direct comparisons between representation spaces, at either the dataset or sample level. At the dataset level, methods such as RSA, SVCCA, and CKA measure global similarity between representation spaces \citep{kriegeskorte2008representational,raghu2017svcca,kornblith2019similarity}. Related studies further examine whether representations learned by different models converge toward similar structures \citep{li2015convergent,Huh2024Platonic}. Such analyses provide an aggregate view of representation similarity, but do not directly capture how cross-model disagreement varies across individual inputs \citep{kolling2025investigating}. Recent sample-level approaches address this limitation by quantifying representational agreement for individual inputs. For example, PNKA compares an input's similarity profiles across two representation spaces, while neighborhood consistency measures the overlap of its nearest-neighbor sets across multiple pretrained models \citep{kolling2025investigating,park2023quantifying}. In essence, prior methods ask a \textbf{measurement} question: given representations, how much do models disagree on a particular input? We instead ask a \textbf{prediction} question: can such \emph{Rashomon Representation}, i.e., cross-model representational disagreement, be predicted from a representation? }

We introduce a supervised framework for sample-level prediction of \emph{rashomon representation} across foundation models. Specifically, we introduce a neighborhood-consistency measure to quantify inter-model representational disagreement. The resulting agreement scores then serve as supervision for training a lightweight predictor that predicts \emph{rashomon representation} directly from the representation produced by a single foundation model. Then, we systematically evaluate the predictability of \emph{rashomon representation} across diverse experimental settings and find that it is consistently predictable from individual model representations. The observed predictability reveals three properties of \emph{rashomon representation}. First, the degree of disagreement varies substantially across different inputs. Second, this variation is not random, but follows systematic and learnable patterns. Third, these patterns can be identified from the representation of a single foundation model, indicating that cross-model disagreement is partially encoded in individual representations. Together, these findings provide a practical mechanism for anticipating representation-level uncertainty at inference time. Our contributions are as follows:
\begin{itemize}[leftmargin=*]
\item We identify and formalize \emph{Rashomon Representation}, the phenomenon that different foundation models may encode the same input in substantially different ways, and formulate a new task of Rashomon Representation prediction to anticipate such cross-model disagreement from the representation of a single model.
\item We develop a practical framework for this task.
Specifically, we quantify sample-level disagreement using cross-model nearest-neighbor consistency over a reference dataset and train a lightweight predictor to estimate this disagreement from representations.
\item Through extensive experiments, we show that representational disagreement is predictable, and generalizable, and further demonstrate its practical utility for efficiently identifying samples that are potentially vulnerable to unreliable representations across foundation models.
\end{itemize}

\section{Methods}

\textcolor{black}{We investigate whether Rashomon Representation, referring to sample-level representational disagreement across foundation models, can be predicted from the representation of a single foundation model. To this end, we formulate its prediction as a supervised learning problem. We first quantify sample-level cross-model representational disagreement by comparing local neighborhoods across the representation spaces of a pool of foundation models, yielding a disagreement score for each sample. These scores are then used as supervision to train a lightweight predictor that predicts agreement scores from representations produced by a designated anchor model. Once trained, the predictor can estimate the expected disagreement of a new sample using only its anchor-model representation at inference time.}

\begin{figure}[htbp]
\centering
\includegraphics[width=\linewidth]{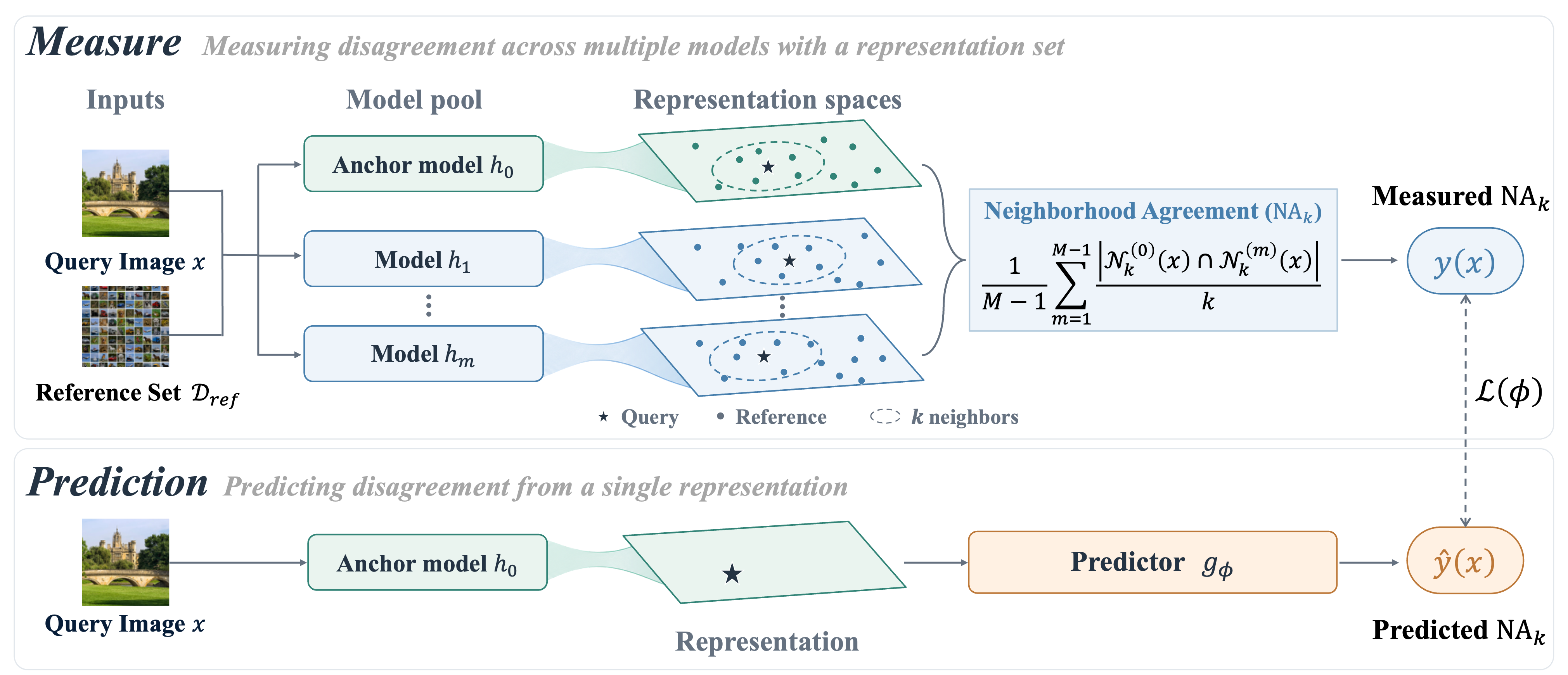}
\caption{Training framework for predicting inter-model representational disagreement. We measure neighborhood agreement between an anchor model and comparison models over a shared reference set to obtain the supervision signal $y(x)$. A lightweight predictor $g_\phi$ learns to approximate $y(x)$ using only the anchor-model representation by minimizing $\mathcal{L}(\phi)$. All foundation models remain frozen, and only the predictor parameters are optimized.}
\label{fig:method}
\end{figure}

\subsection{Quantifying Representational Disagreement}
\textbf{Setup and motivation.}  \textcolor{black}{Given an input sample $x$ and a pool of $M$ pretrained foundation models $\mathcal{H}=\{h_0,h_1,\ldots,h_{M-1}\}$, our goal is to predict, using only the representation $h_0(x)$ produced by an \emph{anchor model} $h_0$, how consistently $x$ is represented across the model pool. We denote the remaining models by $\mathcal{H}_{-0}=\mathcal{H}\setminus\{h_0\}$ and use them to characterize cross-model representational agreement. Directly comparing $h_0(x)$ with $\{h_m(x)\}_{m=1}^{M-1}$ is generally not meaningful, since independently trained models encode samples in model-specific representation spaces that are not directly comparable. Rather than aligning these spaces explicitly, we compare each sample's local neighborhoods across models over the same reference samples, providing a parameter-free comparison that requires neither additional training nor assumptions about cross-model representation alignment. Specifically, for each model, we retrieve the local neighborhood of $x$ from the same reference set and measure cross-model agreement by the overlap between these neighborhoods. This alignment-free consistency score is then used as supervision for our predictor, with lower scores indicating stronger representational disagreement.}

\textbf{Model-specific neighborhood construction.}
\textcolor{black}{Let $\mathcal{D}=\{x_1,\ldots,x_N\}$ denote a reference dataset shared across all models. For each model $h_m$, we characterize how the sample $x$ is locally positioned in the representation space by identifying its $k$ nearest neighbors in $\mathcal{D}$. Our construction only requires a model-specific similarity measure to rank the reference samples and does not depend on a particular choice of metric. In our implementation, we use cosine similarity, a scale-invariant measure commonly used for comparing learned representations. Equivalently, after $\ell_2$ normalization, the similarity between $x$ and a reference sample $x_j \in \mathcal{D}$ is given by
\[
s_m(x,x_j)
=
\frac{
h_m(x)^\top h_m(x_j)
}{
\|h_m(x)\|_2\,\|h_m(x_j)\|_2
}.
\]
We then define $\mathcal{N}_k^{(m)}(x)$ as the index set of the $k$ nearest neighbors with the largest similarity scores:
\[
\mathcal{N}_k^{(m)}(x)
=
\operatorname{TopK}_{j\in\{1,\ldots,N\}}
\, s_m(x,x_j).
\]
Although the resulting neighborhoods are constructed independently in model-specific representation spaces, they are all expressed over the same set of reference indices. Thus, the neighborhoods can be compared directly across models without explicitly aligning their representation spaces.}

\textbf{Cross-model neighborhood agreement.}
\textcolor{black}{Based on model-specific neighborhoods, we quantify how consistently the anchor model $h_0$ represents $x$ relative to the remaining models in $\mathcal{H}_{-0}$. We define
\begin{equation}
\label{eq:na}
\mathrm{NA}_k(x;h_0,\mathcal{H})
=
\frac{1}{M-1}
\sum_{m=1}^{M-1}
\frac{
|
\mathcal{N}_k^{(0)}(x)
\cap
\mathcal{N}_k^{(m)}(x)
|
}{k}.    
\end{equation}
where $|\mathcal{N}_k^{(0)}(x)\cap\mathcal{N}_k^{(m)}(x)|$ denotes the number of $k$-nearest neighbors shared by the anchor model $h_0$ and comparison model $h_m$. $\mathrm{NA}_k$ measures the neighborhood agreement of $x$ at neighborhood size $k$, quantifying how consistently the local neighborhood identified by the anchor model is preserved across the comparison models. For each comparison model $h_m$, the corresponding term measures the fraction of reference neighbors selected by the anchor model that are also selected by $h_m$. Averaging these pairwise overlaps over $\mathcal{H}_{-0}$ yields a sample-level measure of representational consistency relative to $h_0$. The score ranges from $0$ to $1$, with higher values indicating greater agreement in local neighborhood structure across models and lower values indicating stronger representational disagreement.}

\textcolor{black}{\textbf{Neighborhood agreement as supervision.}
Since $\mathrm{NA}_k$ is defined solely through neighborhood overlap, its computation requires no additional model training and does not require other assumptions. It is therefore model-agnostic and can be applied to a broad range of representation models, as long as their representations can be used to construct local neighborhoods. These properties make $\mathrm{NA}_k$ a simple and broadly applicable measure, and we use $\mathrm{NA}_k$ as the sample-level supervision signal for predicting cross-model representational consistency from $h_0(x)$ alone.}
\subsection{From Measurement to Prediction}
\textcolor{black}{\textbf{Predicting cross-model agreement.} The neighborhood agreement score introduced above quantifies, for each sample, the degree of \emph{Rashomon Representation}, i.e., how consistently the same input is represented across different foundation models. However, explicitly computing this score requires access to multiple models and a reference dataset at inference time, limiting its practical applicability. This raises a more fundamental question: can the \emph{Rashomon Representation} be inferred from its representation in a single model? To test this, we use the measured consistency scores as prediction targets and train a lightweight predictor whose only input is the representation produced by an \emph{anchor model}. Successful prediction would indicate that rashomon representation is already reflected, at least partially, in the representation space of an individual model, rather than being a property that becomes observable only after comparing multiple models.}

\textcolor{black}{\textbf{Prediction formulation.} For an input $x$, let $\tilde{z}_0(x)=h_0(x)/\|h_0(x)\|_2\in\mathbb{R}^{d_0}$ denote the $\ell_2$-normalized representation produced by the anchor model, where $d_0$ is its output dimensionality. We introduce a predictor $g_\phi:\mathbb{R}^{d_0}\rightarrow(0,1)$ that produces $\hat{y}(x)=g_\phi(\tilde{z}_0(x))$ to approximate the target $y(x)=\mathrm{NA}_k(x;h_0,\mathcal{H})$. Higher predicted values indicate stronger expected agreement between the anchor and comparison models, while lower values indicate greater representational disagreement. Crucially, the predictor takes only $\tilde{z}_0(x)$ as input, without access to the comparison-model representations or the cross-model neighborhood comparisons used to define the target score. It therefore learns to predict cross-model consistency solely from the representation. If a lightweight predictor can do so reliably, this indicates that cross-model agreement is systematically associated with information already present in the representation.}

\textcolor{black}{\textbf{Predictor construction.}
We instantiate $g_\phi$ as a lightweight residual feed-forward network that maps the normalized anchor representation to a scalar consistency estimate. The input is first projected into a hidden space and then processed by residual blocks composed of LayerNorm, GELU activations, dropout, and identity skip connections before the final prediction layer. Following standard residual feed-forward designs \citep{vaswani2017attention,tolstikhin2021mlp}, we keep the predictor deliberately small and capacity-controlled so that its performance reflects how readily cross-model consistency can be predicted from a single representation, rather than the fitting power of a high-capacity auxiliary model. We additionally evaluate alternative predictor architectures in the experiments to examine whether the predictability depends on a particular architectural choice.}

\textbf{Predictor training.} We use the predefined, disjoint training, validation, and test splits, denoted by $\mathcal{D}_{\mathrm{tr}}$, $\mathcal{D}_{\mathrm{val}}$, and $\mathcal{D}_{\mathrm{te}}$, respectively, and set $\mathcal{D}_{\mathrm{ref}}=\mathcal{D}_{\mathrm{tr}}$. For each sample $x$, we precompute the target $y(x)=\mathrm{NA}_k(x;h_0,\mathcal{H})$ using Eq.~\ref{eq:na}, excluding the query itself for samples in $\mathcal{D}_{\mathrm{tr}}$. The predictor is trained on $\mathcal{D}_{\mathrm{tr}}$ by minimizing 
\[ \mathcal{L}(\phi) = \frac{1}{|\mathcal{D}_{\mathrm{tr}}|} \sum_{x_i\in\mathcal{D}_{\mathrm{tr}}} \ell\!\left( g_\phi(\tilde{z}_0(x_i)), y(x_i) \right),
\] where $\ell$ denotes the regression loss. The checkpoint with the lowest validation loss on $\mathcal{D}_{\mathrm{val}}$ is selected and finally evaluated on $\mathcal{D}_{\mathrm{te}}$. All foundation models
remain fixed, and only the predictor parameters $\phi$ are optimized.

\section{Related Work}
\textbf{Representational Similarity.} A central question in representation analysis is whether different models organize the same inputs in similar ways. Existing methods typically compare the similarity structures induced by different models on a shared dataset. Representational similarity analysis (RSA) measures the correspondence between these structures \citep{kriegeskorte2008representational}, while SVCCA and CKA provide more systematic tools for comparing representations across layers, architectures, and training procedures \citep{raghu2017svcca,kornblith2019similarity}. Related studies further investigate the convergence of independently trained representations and the effects of architectural and pretraining choices \citep{li2015convergent,nguyen2020wide,raghu2021vision,neyshabur2020being,xie2023revealing}. These approaches primarily produce retrospective summaries at the model, layer, or dataset level. More recent work extends representation comparison to individual inputs. For example, PNKA compares an input's similarity profiles across representation spaces, while neighborhood-based methods measure agreement through the overlap of nearest-neighbor sets \citep{park2023quantifying,kolling2025investigating}. Such methods capture input-dependent variation more directly, but remain measurement-oriented: they require representations from multiple models to quantify the disagreement of a given input. In contrast, our work reformulates this problem as prediction. We ask whether cross-model representational disagreement, which we term \emph{Rashomon Representation}, can be inferred from the representation of a single model.

\textbf{Vision Foundation Models} are commonly developed with different pretraining objectives and architectures. Image-language contrastive learning, as used in CLIP, aligns visual representations with natural language \citep{radford2021learning}. Self-supervised methods adopt contrastive, momentum-based, bootstrapping, or self-distillation objectives, including SimCLR, MoCo, BYOL, DINO, and DINOv2 \citep{chen2020simple,he2020momentum,grill2020bootstrap,caron2021emerging,oquab2023dinov2}. Masked-image modeling methods, such as BEiT and MAE, learn representations by recovering masked visual content \citep{bao2021beit,he2022masked}. In terms of architecture, Vision Transformers provide a scalable alternative to convolutional visual encoders \citep{dosovitskiy2020image}. In our experiments, we consider a broader range of vision foundation models. Taking vision foundation models as a representative domain, we conduct extensive experiments to investigate the predictability and generality of cross-model representational disagreement.

\section{Experiments}
Our experiments are designed to test the central hypothesis underlying Rashomon Representation: inter-model representational disagreement is not random, but follows input-dependent patterns that can be predicted from a single model representation. We organize our investigation around two aspects. First, we establish predictability from a single representation and examine whether it persists across model configurations, distribution shifts, and predictor architectures, as well as when the anchor model is excluded from the disagreement target. Second, we assess the practical value of predicted disagreement by examining its association with downstream classification risk and the reductions in computation and reference data required to estimate disagreement.

\textbf{Experimental setup.}
Unless otherwise specified, we use disjoint ImageNet-1K
\citep{deng2009imagenet,russakovsky2015imagenet} splits for predictor
training, validation, and testing, with the full training split as the
neighborhood reference set. Our default configuration uses CLIP ViT-B/32~\citep{radford2021learning}
as the anchor model, with DINOv2 ViT-B/14~\citep{oquab2023dinov2}
and a supervised ViT-B/16 (IN-21K)~\citep{dosovitskiy2020image}
as comparison models. Below, we refer to these models as CLIP,
DINOv2, and ViT-21K, respectively. We use a
lightweight Residual FFN predictor built around residual feed-forward
blocks and set the neighborhood size to $k=50$ when computing the
neighborhood agreement target $\mathrm{NA}_k$. In this default setting,
the predictor estimates neighborhood agreement using only the anchor
representation, with agreement serving as an operational proxy for
inter-model representational disagreement. To address different
experimental questions, we vary model combinations, predictor
architectures, data distributions, reference-set sizes, and other
relevant settings. We report the key settings for each experiment in the
main text and provide additional implementation details in the appendix.

\subsection{Disagreement is predictable}

\begin{figure}[htbp]
\centering
\includegraphics[width=\linewidth]{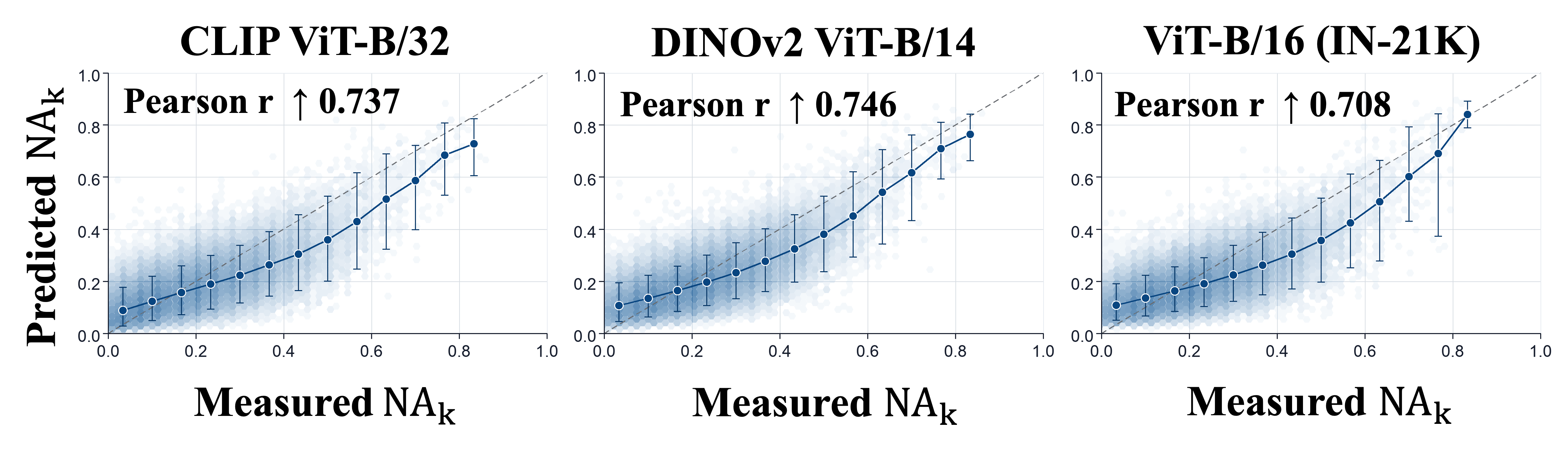}
\vspace{-2\baselineskip}
\caption{Measured and predicted $\mathrm{NA}_k$ across three encoders. Hexagons indicate sample density, while markers and error bars show binned means and 90\% intervals. The strong agreement between measured and predicted values demonstrates that cross-model representational disagreement can be reliably inferred from single-model representations.}
\label{fig:exp1-predictability}
\end{figure}

\paragraph{Disagreement is predictable from a single representation.}
We first ask whether inter-model representational disagreement can be predicted from the representation of a single model. To test this, we take each of CLIP, DINOv2, and ViT-21K in turn as the anchor model and use the other two as comparison models. For each anchor, we train a separate predictor using only the anchor's representation as input to estimate the measured $\mathrm{NA}_k$. As shown in Fig.~\ref{fig:exp1-predictability}, the predicted and measured $\mathrm{NA}_k$ values are positively correlated for all three anchor models, with Pearson correlation coefficients ranging from 0.708 to 0.746. These results support the hypothesis that inter-model representational disagreement leaves a detectable sample-level signature within each model's representation space.

\vspace{-0.5\baselineskip}
\textbf{Predictability across model configurations.}
We next examine whether disagreement predictability depends on the anchor model or the set of comparison models. Building on the preceding experiment, we further introduce Swin-B pretrained on ImageNet-22K \citep{liu2021swin}, SigLIP 2 with a ViT-B/16 image encoder at $224\times224$ resolution \citep{tschannen2025siglip2}, and ConvNeXt-B pretrained on ImageNet-22K \citep{liu2022convnext}. Across 11 configurations, we train a separate predictor using only the designated anchor representation. The dataset setup remains unchanged from the preceding experiment, while we vary the composition and size of the comparison set, as well as the choice of anchor model.

\begin{wraptable}[12]{r}{0.49\textwidth}
\vspace{-1\baselineskip}
\centering
\captionsetup{font=small,skip=4pt}
\caption{Disagreement prediction across anchor-model choices.}
\label{tab:anchor-model-choice}

\footnotesize
\setlength{\tabcolsep}{4pt}
\renewcommand{\arraystretch}{0.95}

\begin{tabular*}{\linewidth}{@{\extracolsep{\fill}}clc@{}}
\toprule
\textbf{Comparison models}
&
\textbf{Anchor}
&
\textbf{Pearson $r\uparrow$}
\\
\midrule

\multirow{3}{*}[-1ex]{
    \shortstack[c]{
        Swin-B\\[-1pt]
        SigLIP 2\\[-1pt]
        ConvNeXt-B
    }
}
&
CLIP
&
0.734
\\
\cmidrule(l){2-3}

&
DINOv2
&
0.751
\\
\cmidrule(l){2-3}

&
ViT-21K
&
0.701
\\

\midrule

\multirow{3}{*}[-2ex]{
    \shortstack[c]{
        CLIP\\[-1pt]
        DINOv2\\[-1pt]
        ViT-21K
    }
}
&
Swin-B
&
0.770
\\
\cmidrule(l){2-3}

&
SigLIP 2
&
0.750
\\
\cmidrule(l){2-3}

&
ConvNeXt-B
&
0.767
\\

\bottomrule
\end{tabular*}
\vspace{-0.4\baselineskip}
\end{wraptable}

\emph{(i) Anchor model choice.}
To assess how the choice of anchor model affects disagreement
predictability, we evaluate six different models. We use Swin-B,
SigLIP 2, and ConvNeXt-B as the comparison model set when evaluating
CLIP, DINOv2, and ViT-21K, and use CLIP, DINOv2, and ViT-21K as the comparison model set when evaluating the other three models.
As shown in Table~\ref{tab:anchor-model-choice}, predicted and measured
$\mathrm{NA}_k$ remain positively correlated across all six
configurations, with Pearson $r$ ranging from 0.701 to 0.770. These
results show that inter-model representational disagreement can be
learned regardless of which model supplies the representation to the
predictor and serves as the anchor model in the agreement computation.

\emph{(ii) Composition of the comparison model set.}
We first fix DINOv2 as the anchor model and the number of models in the comparison model set at three, while varying which model types constitute the set. As shown in Table~\ref{tab:comparison-composition}, the three comparison model sets yield Pearson correlations of 0.751, 0.754, and 0.762, respectively. The consistently high positive correlations indicate that inter-model representational disagreement remains predictable across different compositions of the comparison model set and is not limited to one particular model combination.

\emph{(iii) Comparison model set size.}
We next examine whether inter-model representational disagreement
remains predictable when the number of models in the comparison model
set changes. To test this, we fix CLIP as the anchor model and
use nested comparison model sets containing two to five models. As
shown in Table~\ref{tab:comparison-panel-size}, performance remains comparable
across all four settings, with Pearson $r$ ranging from 0.727 to 0.750.
These results show that inter-model representational disagreement can
be learned across the tested comparison model set sizes and maintains
strong predictive performance when more models are included in the
comparison model set.

\begin{table*}[htbp]
\centering
\captionsetup{font=small,skip=4pt}
\caption{
Disagreement prediction across comparison-model compositions and
comparison model set sizes.
}
\label{tab:comparison-ablation}

\vspace{-0.25em}

\begin{subtable}[t]{0.38\textwidth}
\centering
\captionsetup{font=small,skip=3pt}
\caption{Composition (DINOv2 anchor).}
\label{tab:comparison-composition}

\footnotesize
\renewcommand{\arraystretch}{1.49}

\resizebox{\linewidth}{!}{%
\begin{tabular}{@{}l@{\hspace{0.5em}}r@{}}
\toprule
\textbf{Comparison models}
&
\textbf{Pearson $r\uparrow$}
\\
\midrule

Swin-B, SigLIP 2, ConvNeXt-B
&
0.751
\\

Swin-B, SigLIP 2, ViT-21K
&
0.754
\\

CLIP, SigLIP 2, ConvNeXt-B
&
0.762
\\

\bottomrule
\end{tabular}%
}

\end{subtable}
\hfill
%
\begin{subtable}[t]{0.60\textwidth}
\centering
\captionsetup{font=small,skip=3pt}
\caption{Set size (CLIP anchor).}
\label{tab:comparison-panel-size}

\footnotesize
\renewcommand{\arraystretch}{1.045}

\resizebox{\linewidth}{!}{%
\begin{tabular}{@{}l@{\hspace{0.3em}}r@{}}
\toprule
\textbf{Comparison models}
&
\textbf{Pearson $r\uparrow$}
\\
\midrule

Swin-B, SigLIP 2
&
0.727
\\

Swin-B, SigLIP 2, ConvNeXt-B
&
0.734
\\

Swin-B, SigLIP 2, ConvNeXt-B, DINOv2
&
0.744
\\

Swin-B, SigLIP 2, ConvNeXt-B, DINOv2, ViT-21K
&
0.750
\\

\bottomrule
\end{tabular}%
}

\end{subtable}

\vspace{-0.35em}
\end{table*}

Taken together, these experiments show that the predictability of inter-model representational disagreement persists across comparison panel compositions, anchor model choices, and panel sizes. These variations modify complementary components of the protocol, including the models used to construct the agreement target, the predictor input, and the breadth of the comparison panel. Nevertheless, the positive association between predicted and measured $\mathrm{NA}_k$ remains consistent, indicating that the observed predictability is not confined to any single model combination.

\paragraph{Disagreement remains predictable under distribution shift.}
We next ask whether disagreement remains predictable when the query distribution changes. To test this, we evaluate the ImageNet-1K-trained predictor on nine external datasets, four natural ImageNet shifts, and five Gaussian-noise levels without retraining or test-time adaptation. For each query, predicted $\mathrm{NA}_k$ is produced by applying this fixed predictor to the query's CLIP representation, while measured $\mathrm{NA}_k$ is computed by comparing the neighborhoods retrieved by CLIP, DINOv2, and ViT-21K from the same fixed ImageNet-1K training reference set. As shown in Tables~\ref{tab:exp3-cross-dataset} and~\ref{tab:exp3-imagenet-variants}, predicted and measured $\mathrm{NA}_k$ remain positively correlated across all settings, with Pearson $r$ ranging from 0.405 to 0.661 on the external datasets, from 0.710 to 0.720 on ImageNet-V2, and reaching 0.512 on ImageNet-R. These results show that disagreement predictability transfers beyond the training distribution, although its strength varies across domains. This setting reflects practical use: the data available during training are finite, while future queries may come from a much broader range of distributions.

\vspace{-1\baselineskip}

\begin{table}[htbp]
\caption{Cross-dataset disagreement prediction. Pearson correlations between predicted and measured $\mathrm{NA}_k$ on nine external datasets using the fixed ImageNet-1K-trained predictor. Dataset sources: Aircraft \citep{maji2013fgvcaircraft}, Caltech101 \citep{feifei2004caltech101}, Cars \citep{krause2013stanfordcars}, DTD \citep{cimpoi2014dtd}, EuroSAT \citep{helber2019eurosat}, Food101 \citep{bossard2014food101}, Pets \citep{parkhi2012catsdogs}, SUN397 \citep{xiao2010sun}, and UCF101 \citep{soomro2012ucf101}.}
\label{tab:exp3-cross-dataset}
\centering

\begingroup
\small
\renewcommand{\arraystretch}{1.10}
\setlength{\tabcolsep}{2.0pt}

\begin{tabular*}{\linewidth}{
@{\extracolsep{\fill}}
lccccccccc
@{}
}
\toprule

&
Aircraft
& Caltech101
& Cars
& DTD
& EuroSAT
& Food101
& Pets
& SUN397
& UCF101
\\

\midrule

Pearson $r\uparrow$
& 0.661
& 0.656
& 0.425
& 0.587
& 0.405
& 0.585
& 0.499
& 0.597
& 0.637
\\

\bottomrule
\end{tabular*}

\endgroup
\end{table}

\vspace{-1\baselineskip}

\begin{table}[htbp]
\caption{Disagreement prediction under ImageNet distribution shifts. Pearson correlations between predicted and measured $\mathrm{NA}_k$ on ImageNet-V2 \citep{recht2019imagenet}, ImageNet-R \citep{hendrycks2021many}, and Gaussian-corrupted ImageNet-C \citep{hendrycks2019robustness}. The labels S1, S2, S3, S4, and S5 denote corruption severity levels 1, 2, 3, 4, and 5, respectively.}
\label{tab:exp3-imagenet-variants}
\centering

\begingroup
\small
\renewcommand{\arraystretch}{1.10}
\setlength{\tabcolsep}{1.8pt}

\begin{tabular*}{\linewidth}{
@{\extracolsep{\fill}}
lccccccccc
@{}
}
\toprule

&
\multicolumn{3}{c}{ImageNet-V2}
&
\multirow{2}{*}{ImageNet-R}
&
\multicolumn{5}{c}{ImageNet-C: Gaussian}
\\

\cmidrule(lr){2-4}
\cmidrule(lr){6-10}

&
Matched
& Threshold 0.7
& Top Images
&
&
S1
& S2
& S3
& S4
& S5
\\

\midrule

Pearson $r\uparrow$
&
0.710
& 0.720
& 0.717
&
0.512
&
0.726
& 0.732
& 0.724
& 0.650
& 0.394
\\

\bottomrule
\end{tabular*}

\endgroup
\end{table}

\begin{figure}[htbp]
    \centering
    \includegraphics[width=\linewidth]{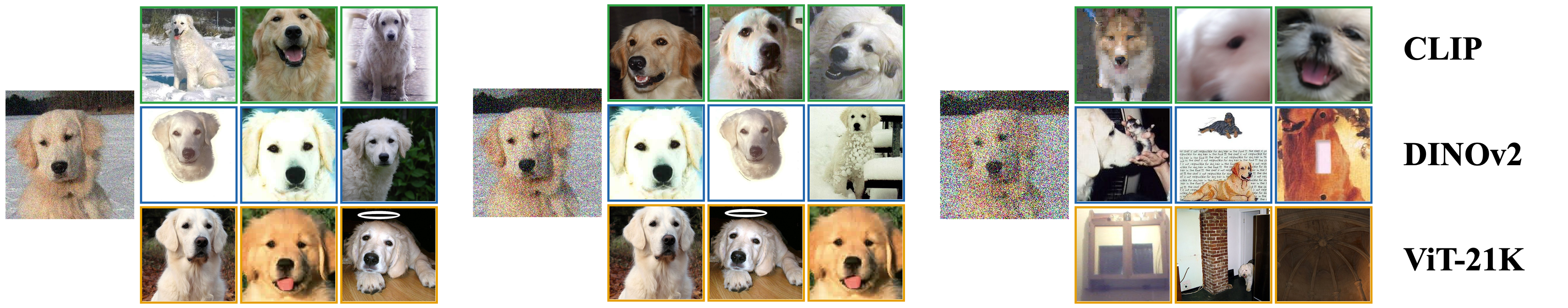}
    \caption{Prediction drift under Gaussian-noise corruption. From left to right, corruption severity increases; each example shows a query and its three nearest neighbors retrieved by CLIP, DINOv2, and ViT-21K. }
    \label{fig:exp3-corruption-cases}
\end{figure}

Under Gaussian noise, Pearson $r$ decreases from 0.726 at severity 1 to 0.394 at severity 5. In the three examples shown in Figure~\ref{fig:exp3-corruption-cases}, absolute errors increase from 0.013 to 0.031 and 0.037 as the neighborhoods retrieved by CLIP, DINOv2, and ViT-21K become increasingly divergent. These observations suggest that severe corruption pushes query representations beyond the regime learned from ImageNet-1K, causing the learned mapping from anchor representations to cross-model agreement to become unreliable and the predictions to drift.

\begin{wrapfigure}[11]{r}{0.48\textwidth}
\vspace{-1\baselineskip}
\centering
\includegraphics[width=\linewidth]{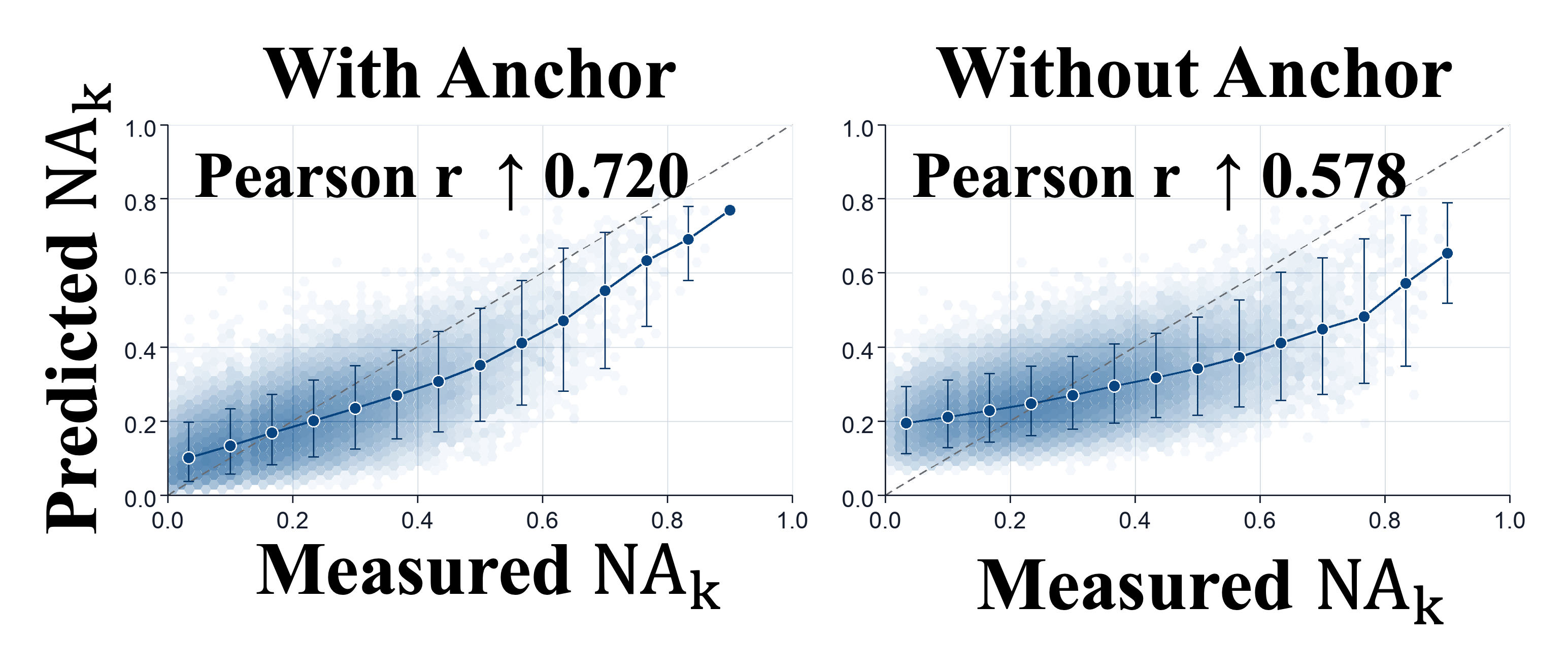}
\captionsetup{font=small,skip=2pt}
\caption{Leave-anchor-out diagnosis on ImageNet-1K. Both panels use only CLIP as input; the target includes CLIP on the left and excludes it on the right.}
\label{fig:leave-anchor-out}
\vspace{-1.2\baselineskip}
\end{wrapfigure}

\textbf{Predictability beyond the anchor.}
In the preceding experiments, the anchor model provides the predictor input, and its neighborhood also contributes to the supervision target.
We further ask whether a single model's representation can predict representational disagreement within a specified set of other models.
To test this, we train a predictor using only CLIP representations, with the target defined as the average pairwise top-$50$ neighborhood agreement among DINOv2, ViT-21K, and Swin-B, excluding CLIP.
As Figure~\ref{fig:leave-anchor-out} shows, the predicted and measured agreement remain correlated, with Pearson $r = 0.578$.
For this model set, the result shows that a CLIP representation provides information for predicting disagreement among models that do not include CLIP.

\begin{wraptable}[11]{r}{0.48\textwidth}
\vspace{-0.8\baselineskip}
\centering
\captionsetup{font=small,skip=4pt,width=\linewidth}
\caption{Predictor architecture comparison measured by Pearson correlation. All predictors use the same single-representation input and measured consistency supervision.}
\label{tab:predictor-architecture}
\begingroup
\footnotesize
\renewcommand{\arraystretch}{1}
\begin{tabular*}{\linewidth}{@{}l@{\extracolsep{\fill}}ccc@{}}
\toprule
\textbf{Model} &
\shortstack{\textbf{2-layer}\\\textbf{MLP}} &
\shortstack{\textbf{Transformer}\\\textbf{Encoder}} &
\shortstack{\textbf{Residual}\\\textbf{FFN}} \\
\midrule
CLIP & 0.544 & 0.712 & \textbf{0.737} \\
DINOv2 & 0.523 & 0.724 & \textbf{0.746} \\
ViT-21K & 0.490 & 0.696 & \textbf{0.708} \\
\bottomrule
\end{tabular*}
\endgroup
\end{wraptable}
\textbf{Predictability holds across predictor architectures.} We assess architectural sensitivity by comparing a 2-layer MLP, Residual FFN, and Transformer Encoder across three anchor models, holding the predictor input and measured $\mathrm{NA}_k$ target fixed across architectures for each anchor. As shown in Table~\ref{tab:predictor-architecture}, all three architectures yield positive correlations for every anchor; the 2-layer MLP achieves Pearson $r$ values ranging from 0.490 to 0.544, while the Residual FFN reaches from 0.708 to 0.746. These results show that  different architectures recover the disagreement signal, indicating that its predictability is not an artifact of a particular design, although architecture affects prediction accuracy.

\subsection{Predicted disagreement is practical}

\begin{wraptable}[9]{r}{0.59\textwidth}
\vspace{-1\baselineskip}
\centering
\captionsetup{font=small,skip=4pt}
\caption{Post-embedding cost on 100K ImageNet-1K samples. Lower is better.}
\label{tab:exp7-cost-benchmark}

\small
\setlength{\tabcolsep}{3.5pt}
\renewcommand{\arraystretch}{1.05}

\begin{tabular*}{\linewidth}{@{\extracolsep{\fill}}lccc@{}}
\toprule
Method
& \shortstack{Time $\downarrow$\\[-1pt]\scriptsize(ms/sample)}
& \shortstack{Peak memory $\downarrow$\\[-1pt]\scriptsize(MiB)}
& \shortstack{Storage $\downarrow$\\[-1pt]\scriptsize(MiB)}
\\
\midrule
Measured (PyTorch) & 49.54 & 20,643 & 10,009 \\
Measured (FAISS)   & 20.54 & 29,729 & 10,009 \\
\midrule
\textbf{Predicted}
& \textbf{5.34}
& \textbf{9}
& \textbf{5}
\\
\bottomrule
\end{tabular*}
\end{wraptable}

\paragraph{Prediction substantially reduces measurement cost.}
Computing measured $\mathrm{NA}_k$ requires neighborhood search across multiple representation spaces for every query. We compare its PyTorch and FAISS implementations with predicted $\mathrm{NA}_k$, obtained through a single forward pass over the CLIP representation, on $100{,}000$ ImageNet-1K test samples. As shown in Table~\ref{tab:exp7-cost-benchmark}, predicted $\mathrm{NA}_k$ takes only 5.34 ms per sample, making it $3.85\times$ faster than FAISS and $9.28\times$ faster than PyTorch, while reducing GPU memory and asset storage by over three orders of magnitude. These results show that prediction provides an efficient alternative to repeatedly measuring inter-model representational disagreement.

\textbf{Predicted disagreement tracks downstream classification risk.}
Inter-model representational disagreement reflects differences in how foundation models organize the same input, which may lead to different behavior on downstream tasks. We therefore ask whether predicted disagreement retains the association between measured disagreement and downstream classification risk. Across nine benchmarks, we compute Kendall $\tau_b$ correlations of measured and predicted disagreement, each defined as $1-\mathrm{NA}_k$, with the per-sample multiclass Brier risk of fixed, dataset-specific CLIP linear probes. As shown in Table~\ref{tab:exp6-downstream-risk-pearson}, predicted disagreement is positively associated with risk on every dataset, and the average $\tau_b$ increases from 0.18 for measured disagreement to 0.21 for predicted disagreement. One possible explanation is that training filters some noise in the measured $\mathrm{NA}_k$ targets, allowing predicted disagreement to align more closely with downstream risk. These results indicate that predicted disagreement retains an association with downstream classification risk.
\begin{table*}[htbp]
\caption{Predicted representational disagreement tracks downstream classification risk. We report sample-wise Kendall $\tau_b$ correlations between measured or predicted disagreement and multiclass Brier risk across nine benchmarks. Red values indicate increases over directly measured disagreement; green values indicate decreases. }
\label{tab:exp6-downstream-risk-pearson}
\centering

\begingroup
\normalsize
\setlength{\tabcolsep}{3.4pt}
\renewcommand{\arraystretch}{1.12}

\newcommand{\tablehead}[1]{{\small\bfseries #1}}

\newcommand{\increasecell}[2]{%
  \shortstack{%
    #1\\[-1pt]
    {\small\color[RGB]{190,45,45}#2}%
  }%
}

\newcommand{\decreasecell}[2]{%
  \shortstack{%
    #1\\[-1pt]
    {\small\color[RGB]{34,139,83}#2}%
  }%
}

\newcommand{\avgincreasecell}[2]{%
  \shortstack{%
    \textbf{#1}\\[-1pt]
    {\small\bfseries\color[RGB]{190,45,45}#2}%
  }%
}

\newcommand{\predictedlabel}{%
  \shortstack[l]{%
    \textbf{Predicted}\\[-1pt]
    {\small\phantom{+0.00}}%
  }%
}

\resizebox{\textwidth}{!}{%
\begin{tabular}{@{}l*{10}{c}@{}}
\toprule
&
\tablehead{Aircraft}
&
\tablehead{Caltech101}
&
\tablehead{Cars}
&
\tablehead{DTD}
&
\tablehead{EuroSAT}
&
\tablehead{Food101}
&
\tablehead{Pets}
&
\tablehead{SUN397}
&
\tablehead{UCF-101}
&
\tablehead{Average}
\\
\midrule

\textbf{Measured}
& 0.24
& 0.20
& 0.09
& 0.16
& 0.09
& 0.32
& 0.16
& 0.20
& 0.18
& \textbf{0.18}
\\

\midrule

\predictedlabel
& \increasecell{0.26}{+0.02}
& \increasecell{0.23}{+0.03}
& \increasecell{0.10}{+0.01}
& \increasecell{0.17}{+0.01}
& \decreasecell{0.08}{-0.01}
& \increasecell{0.40}{+0.08}
& \increasecell{0.17}{+0.01}
& \increasecell{0.23}{+0.03}
& \increasecell{0.23}{+0.05}
& \avgincreasecell{0.21}{+0.03}
\\

\bottomrule
\end{tabular}%
}

\endgroup
\end{table*}

\begin{wrapfigure}[16]{r}{0.45\textwidth}
\vspace{-0.5\baselineskip}
\centering
\includegraphics[width=\linewidth]{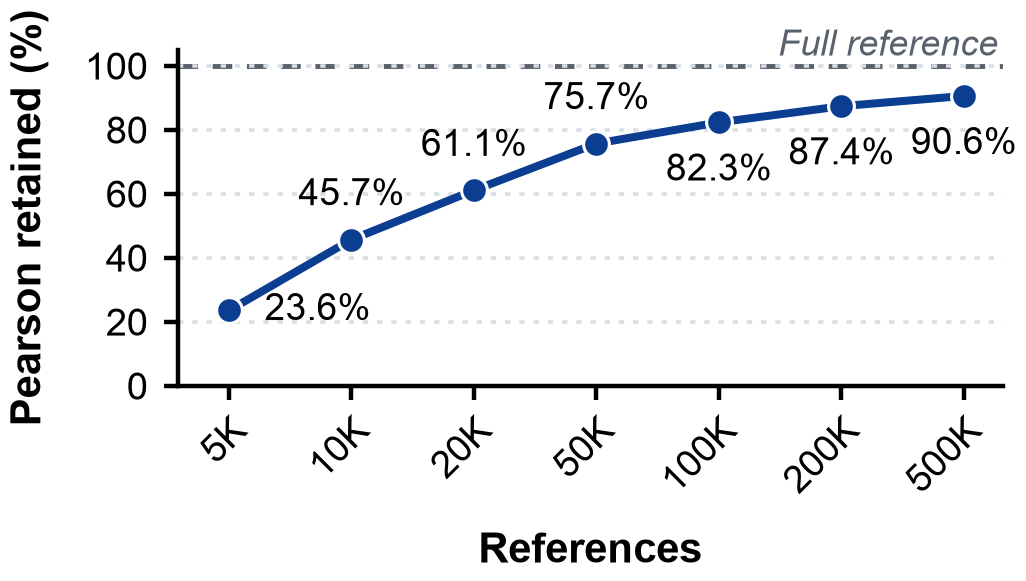}
\captionsetup{font=small,skip=3pt}
\caption{Reference set scaling for disagreement prediction. Pearson correlation retained relative to using the full ImageNet-1K training set as the reference set; all predictors use the same 50,000 training samples and are evaluated against the same full-reference $\mathrm{NA}_k$ target.}
\label{fig:reference-scaling}
\vspace{-0.45\baselineskip}
\end{wrapfigure}

\textbf{Prediction reduces data requirements.} Obtaining measured $\mathrm{NA}_k$ typically requires a large reference set. In contrast, estimating inter-model disagreement through prediction can leverage learned generalization to reduce dependence on reference-set size. We construct nested reference subsets ranging from 5K to 500K samples and use the measured $\mathrm{NA}_k$ from each subset to train a predictor on the same fixed 50,000 training samples. We then compute the Pearson correlation between predicted $\mathrm{NA}_k$ and measured $\mathrm{NA}_k$ obtained offline using the full reference set. To quantify the retained ability to approximate full-reference measured $\mathrm{NA}_k$, we divide this correlation by that achieved by a predictor trained with full-reference supervision. As shown in Figure~\ref{fig:reference-scaling}, even 100K references retain 82.3\% of the full-reference predictive correlation, while 500K references, fewer than half of the full reference set, retain 90.6\% and approach full-reference performance with substantially less data.

\section{Conclusion}
We presented a framework for predicting \emph{Rashomon Representation}, the sample-dependent disagreement among foundation models, from a single model's representation. Cross-model neighborhood agreement operationalizes this disagreement and provides supervision for a lightweight predictor. Across diverse vision foundation models and evaluation settings, predicted agreement consistently tracks measured agreement, indicating that inter-model representational disagreement is structured, input-dependent, and partially encoded in individual representation spaces. This enables efficient inference-time estimation without computing representations and neighborhoods for every comparison model. Future work will investigate the mechanisms underlying this predictability and develop self-supervised approaches that learn representational disagreement without explicit cross-model measurements.

\bibliography{references}
\bibliographystyle{plainnat}

\appendix
\newpage
\section{Common Experimental Setup}
\label{app:common-setup}

\subsection{Data Splits}

We use predefined ImageNet-1K splits for predictor training, model
selection, and final evaluation. The training, validation, and test
splits, denoted by $\mathcal{D}_{\mathrm{tr}}$,
$\mathcal{D}_{\mathrm{val}}$, and $\mathcal{D}_{\mathrm{te}}$, contain
1,281,167, 50,000, and 100,000 samples, respectively.

Except in the reference-set scaling experiment, the full training split
serves as the shared neighborhood reference set:
\begin{equation}
\mathcal{D}_{\mathrm{ref}}=\mathcal{D}_{\mathrm{tr}}.
\end{equation}
Neighborhoods for training, validation, and test queries are constructed
with respect to this reference set. Representations and neighborhood
sets are aligned by image identity across models so that cross-model
comparisons refer to the same reference samples.

\subsection{Foundation Models and Representations}

We use six frozen vision encoders. Each model follows its corresponding
image preprocessing pipeline, and the extracted representations are used
for neighborhood construction and predictor training.
Table~\ref{tab:app-backbones} summarizes the model configurations.

\begin{table}[htbp]
\centering
\caption{Foundation models and extracted representations.}
\label{tab:app-backbones}
\begingroup
\small
\setlength{\tabcolsep}{4pt}
\renewcommand{\arraystretch}{1.15}
\resizebox{\linewidth}{!}{%
\begin{tabular}{llll}
\toprule
Model & Model source & Representation & Dimension \\
\midrule
CLIP ViT-B/32
& \texttt{openai/clip-vit-base-patch32}
& Projected image features
& 512 \\
DINOv2 ViT-B/14
& \texttt{facebook/dinov2-base}
& Pooled output; CLS if unavailable
& 768 \\
ViT-B/16
& \texttt{google/vit-base-patch16-224-in21k}
& Pooled output; CLS if unavailable
& 768 \\
Swin-B
& \texttt{microsoft/swin-base-patch4-window7-224-in22k}
& Pooled output
& 1024 \\
SigLIP 2 ViT-B/16
& \texttt{google/siglip2-base-patch16-224}
& Projected image features
& 768 \\
ConvNeXt-B
& \texttt{timm/convnext\_base.fb\_in22k}
& Pooled pre-logit features
& 1024 \\
\bottomrule
\end{tabular}%
}
\endgroup
\end{table}

The normalized representation of sample $x$ under model $h_m$ is
\begin{equation}
\tilde{z}_m(x)
=
\frac{h_m(x)}{\|h_m(x)\|_2}.
\end{equation}
All foundation models remain frozen throughout the experiments.
Cross-model neighborhood comparisons operate on reference image
identities and do not require equal representation dimensionality.

\subsection{Neighborhood Construction and Supervision}

Following the main text, let
\begin{equation}
\mathcal{H}=\{h_0,h_1,\ldots,h_{M-1}\}
\end{equation}
denote the foundation-model pool, where $h_0$ is the anchor and the
remaining models form the comparison set. The similarity between query
$x$ and reference sample $x_j$ under model $h_m$ is
\begin{equation}
s_m(x,x_j)
=
\tilde{z}_m(x)^\top\tilde{z}_m(x_j).
\end{equation}

We perform exact nearest-neighbor search using FAISS FlatIP to obtain
$\mathcal{N}_k^{(m)}(x)$, with $k=50$ by default. Pairwise neighborhood
agreement is
\begin{equation}
\label{eq:app-pairwise-agreement}
a_k^{(u,v)}(x)
=
\frac{
\left|
\mathcal{N}_k^{(u)}(x)
\cap
\mathcal{N}_k^{(v)}(x)
\right|
}{k}.
\end{equation}
The default supervision target is
\begin{equation}
\label{eq:app-agreement-target}
y(x)
=
\mathrm{NA}_k(x;h_0,\mathcal{H})
=
\frac{1}{M-1}
\sum_{m=1}^{M-1}a_k^{(0,m)}(x).
\end{equation}

This target lies in $[0,1]$. Higher values indicate greater agreement
between the local neighborhood structures induced by the anchor and
comparison models, corresponding to weaker inter-model
representational disagreement.

\subsection{Predictor Training}

The predictor uses only the anchor representation:
\begin{equation}
\hat{y}(x)
=
g_\phi\!\left(\tilde{z}_0(x)\right).
\end{equation}
Comparison-model representations and neighborhoods are used to
construct offline supervision and do not enter the predictor's forward
pass. Except in the architecture comparison, we use the Residual FFN
described in Appendix~\ref{app:architectures}.

The predictor minimizes
\begin{equation}
\mathcal{L}(\phi)
=
\frac{1}{|\mathcal{D}_{\mathrm{tr}}|}
\sum_{x_i\in\mathcal{D}_{\mathrm{tr}}}
\ell\!\left(
g_\phi(\tilde{z}_0(x_i)),
y(x_i)
\right).
\end{equation}
We use a mean squared error (MSE) objective with a fixed scaling
factor of $1/2$. The per-sample loss is
\begin{equation}
\ell(\hat{y},y)
=
\frac{1}{2}(\hat{y}-y)^2.
\end{equation}

Unless otherwise specified, we use AdamW with a learning rate and
weight decay of $10^{-4}$ and train for 10 epochs. Checkpoints are
selected by validation MAE. The selected parameters, denoted by
$\phi^\star$, are evaluated on the test split. Only predictor
parameters are optimized.

\section{Common Evaluation and Visualization Procedures}
\label{app:evaluation}

\subsection{Pearson Correlation}

For two nonconstant sample sequences $a=(a_i)_{i=1}^{n}$ and
$b=(b_i)_{i=1}^{n}$, Pearson correlation is
\begin{equation}
\label{eq:app-pearson}
\operatorname{Corr}_{\mathrm{P}}(a,b)
=
\frac{
\sum_{i=1}^{n}(a_i-\bar{a})(b_i-\bar{b})
}{
\sqrt{\sum_{i=1}^{n}(a_i-\bar{a})^2}
\sqrt{\sum_{i=1}^{n}(b_i-\bar{b})^2}
},
\end{equation}
where
\begin{equation}
\bar{a}=\frac{1}{n}\sum_{i=1}^{n}a_i,
\qquad
\bar{b}=\frac{1}{n}\sum_{i=1}^{n}b_i.
\end{equation}

In the disagreement prediction experiments, let
\begin{equation}
y_i=y(x_i),
\qquad
\hat{y}_i=g_{\phi^\star}(\tilde{z}_0(x_i)).
\end{equation}
The reported correlation is
\begin{equation}
r_{\mathrm{pred}}
=
\operatorname{Corr}_{\mathrm{P}}(\hat{y},y),
\end{equation}
which measures the linear association between predicted and measured
neighborhood agreement. 

\subsection{Binned Means and Distribution Intervals}

Sample-level figures display the two-dimensional density of
$(y_i,\hat{y}_i)$ and group samples into bins of measured agreement.
For a nonempty bin $I_b$, define
\begin{equation}
\mathcal{I}_b=\{i:y_i\in I_b\}.
\end{equation}
The mean measured and predicted agreement within the bin are
\begin{equation}
\bar{y}_b
=
\frac{1}{|\mathcal{I}_b|}
\sum_{i\in\mathcal{I}_b}y_i,
\qquad
\overline{\hat{y}}_b
=
\frac{1}{|\mathcal{I}_b|}
\sum_{i\in\mathcal{I}_b}\hat{y}_i.
\end{equation}

The binned means summarize prediction trends across measured agreement
levels. Vertical intervals extend from the 5th to the 95th percentile
of predictions within each bin, representing the central 90\% of the
prediction distribution.

\section{Qualitative Visualization of Inter-Model Disagreement}
\label{app:qualitative-disagreement}

\begin{figure*}[htbp]
    \centering
    \includegraphics[width=\textwidth]{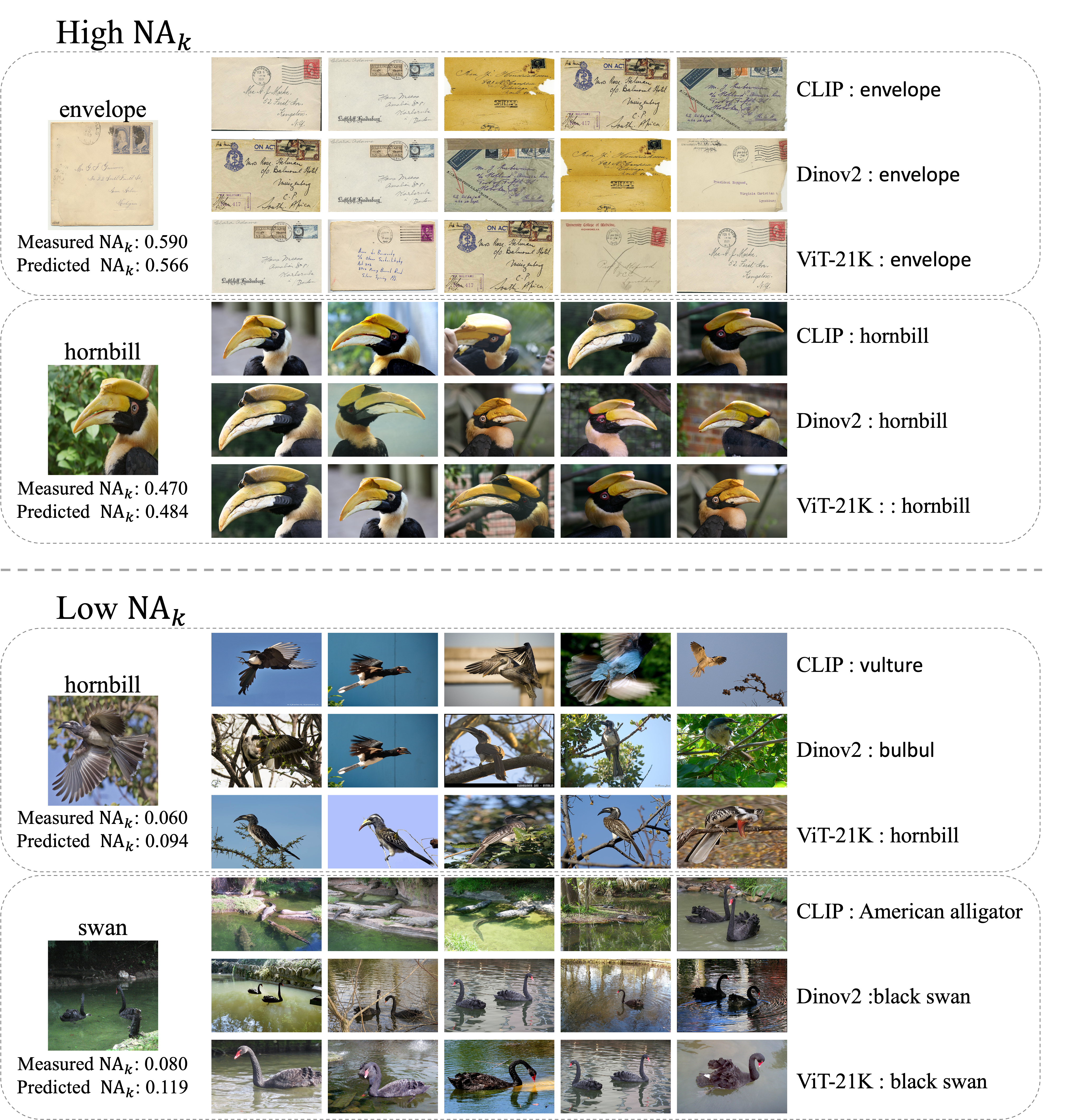}
    \caption{High (top) and low (bottom) cross-model neighborhood agreement. Rows correspond to CLIP, DINOv2, and ViT-21K.}
    \label{fig:app-qualitative-na}
\end{figure*}

Figure~\ref{fig:app-qualitative-na} illustrates
\emph{Rashomon Representation} through high- and low-agreement
queries. For the envelope and hornbill examples, CLIP, DINOv2, and
ViT-21K retrieve similar neighbors, and their linear probes predict
the correct class. For the flying-bird and black-swan examples, the
neighborhoods diverge and the probes assign different classes. The
Residual FFN, using only the CLIP representation, preserves this
high--low ordering in its predicted $\mathrm{NA}_{50}$.

Across ImageNet-1K and ImageNet-V2 variants, we use the ImageNet-1K
training set as the shared reference set. Measured $\mathrm{NA}_{50}$
averages CLIP--DINOv2 and CLIP--ViT-21K Top-50 overlap; the figure
shows only Top-5 neighbors. In the 10,000-query case-study cohort,
all three probes are correct for 24.1\% of queries with
$\mathrm{NA}_{50}<0.10$, versus 68.3\% with
$\mathrm{NA}_{50}\geq0.40$, indicating an association between
neighborhood organization and classification outcomes.

\section{Predicting Disagreement from a Single Representation}
\label{app:single-representation}

This experiment tests whether inter-model disagreement can be predicted
from a single model representation. We separately use CLIP, DINOv2,
and ViT as anchors, with the other two models defining the supervision
target.

\begin{table}[htbp]
\centering
\caption{Configurations for prediction from a single representation.}
\label{tab:app-base-configurations}
\begin{tabular}{llc}
\toprule
Anchor model & Comparison models & Input dimension \\
\midrule
CLIP   & DINOv2, ViT-21K & 512 \\
DINOv2 & CLIP, ViT-21K  & 768 \\
ViT-21K    & CLIP, DINOv2 & 768 \\
\bottomrule
\end{tabular}
\end{table}

A corresponding Residual FFN is trained for each configuration using
only the designated anchor representation. All three configurations
share the data splits, reference set, neighborhood size, and training
protocol. We evaluate the association between predicted and measured
agreement on the ImageNet-1K test split.

\section{Effect of Comparison-Model Composition}
\label{app:comparison-composition}

This experiment fixes DINOv2 as the anchor and uses three comparison
models while varying their composition.

\begin{table}[htbp]
\centering
\caption{Comparison-model compositions with DINOv2 fixed as the anchor.}
\label{tab:app-comparison-composition}
\begin{tabular}{cl}
\toprule
Configuration & Comparison models \\
\midrule
1 & Swin, SigLIP 2, ConvNeXt \\
2 & Swin, SigLIP 2, ViT-21K \\
3 & CLIP, SigLIP 2, ConvNeXt \\
\bottomrule
\end{tabular}
\end{table}

All configurations use the 768-dimensional DINOv2 representation as
predictor input. We construct a separate supervision target for each
comparison set and train a corresponding Residual FFN. The data splits,
reference set, $k=50$, and training protocol remain fixed.

\section{Effect of Anchor-Model Choice}
\label{app:anchor-choice}

This experiment examines disagreement predictability across different
anchor representations. Each configuration uses three comparison
models.

\begin{table}[htbp]
\centering
\caption{Configurations for evaluating anchor-model choice.}
\label{tab:app-anchor-choice}
\begin{tabular}{ll}
\toprule
Anchor model & Comparison models \\
\midrule
CLIP     & Swin, SigLIP 2, ConvNeXt \\
DINOv2   & Swin, SigLIP 2, ConvNeXt \\
ViT-21K      & Swin, SigLIP 2, ConvNeXt \\
Swin     & DINOv2, ViT-21K, CLIP \\
SigLIP 2 & DINOv2, ViT-21K, CLIP \\
ConvNeXt & DINOv2, ViT-21K, CLIP \\
\bottomrule
\end{tabular}
\end{table}

Each predictor receives only its corresponding anchor representation.
Input dimensionality varies with the anchor, while the hidden dimension
remains 512. A Residual FFN is trained for each configuration using the
same data splits, reference set, neighborhood size, and training
protocol.

\section{Effect of Comparison-Set Size}
\label{app:comparison-size}

This experiment fixes CLIP as the anchor and uses nested comparison
sets containing two to five models.

\begin{table}[htbp]
\centering
\caption{Nested comparison sets with CLIP fixed as the anchor.}
\label{tab:app-comparison-size}
\begin{tabular}{cl}
\toprule
Number of models & Comparison set \\
\midrule
2 & Swin, SigLIP 2 \\
3 & Swin, SigLIP 2, ConvNeXt \\
4 & Swin, SigLIP 2, ConvNeXt, DINOv2 \\
5 & Swin, SigLIP 2, ConvNeXt, DINOv2, ViT-21K \\
\bottomrule
\end{tabular}
\end{table}

For each set, the target averages pairwise neighborhood agreement
between CLIP and each comparison model. All settings use the same CLIP
input representations, reference set, and $k=50$, with a separate
Residual FFN trained for each target. This evaluates prediction
performance as the comparison set is progressively expanded.

\section{Generalization under Distribution Shift}
\label{app:distribution-shift}

This experiment reuses the CLIP-anchor Residual FFN trained on
ImageNet-1K without retraining or test-time adaptation. Measured
agreement is computed using CLIP, DINOv2, and ViT with respect to the
full ImageNet-1K training reference set.

The evaluation datasets match the tables in the main text:

\begin{itemize}[leftmargin=*]
    \item \textbf{External datasets:}
    Aircraft, Caltech101, Cars, DTD, EuroSAT, Food101,
    Oxford-IIIT Pets, SUN397, and UCF101.

    \item \textbf{Natural distribution shifts:}
    ImageNet-R and the ImageNet-V2 Matched Frequency,
    Threshold 0.7, and Top Images variants.

    \item \textbf{Image corruption:}
    ImageNet-C Gaussian Noise at severity levels 1 through 5.
\end{itemize}

The nine external datasets use their corresponding validation splits.
Pearson correlation between predicted and measured agreement is
computed separately for each dataset and corruption severity. The
predictor and reference set remain fixed across evaluation settings.

\section{Predictor Architecture Comparison}
\label{app:architectures}

\subsection{Experimental Setup}

We evaluate the 2-layer MLP, Residual FFN, and Transformer Encoder under
the three anchor/comparison configurations in
Appendix~\ref{app:single-representation}. For a given anchor, all
architectures use the same input representations, supervision targets,
data splits, and training protocol. Each architecture produces a scalar
agreement prediction through a sigmoid output.

Below, $\operatorname{Linear}_{a\to b}$ denotes a learnable affine map
from $a$ to $b$ dimensions, $\operatorname{LN}$ denotes LayerNorm, and
$\operatorname{Drop}_{0.1}$ denotes dropout with probability 0.1. Maps
and normalization layers at different locations have independent
parameters. Dropout is enabled only during training.

\subsection{2-layer MLP}

The 2-layer MLP maps the anchor representation to a 512-dimensional
hidden vector and then to a scalar:
\begin{equation}
u
=
\operatorname{Drop}_{0.1}
\left(
\operatorname{GELU}
\left(
\operatorname{Linear}_{d_0\to512}(\tilde{z}_0(x))
\right)
\right),
\end{equation}
\begin{equation}
\hat{y}(x)
=
\sigma\!\left(
\operatorname{Linear}_{512\to1}(u)
\right),
\end{equation}
where $\sigma$ denotes the sigmoid function.

\subsection{Residual FFN}

The input projection is
\begin{equation}
u^{(0)}
=
\operatorname{Drop}_{0.1}
\left(
\operatorname{GELU}
\left(
\operatorname{LN}
\left(
\operatorname{Linear}_{d_0\to512}(\tilde{z}_0(x))
\right)
\right)
\right).
\end{equation}

Two residual feed-forward blocks are then applied:
\begin{equation}
u^{(\ell)}
=
u^{(\ell-1)}
+
F_\ell
\left(
\operatorname{LN}_\ell(u^{(\ell-1)})
\right),
\qquad \ell=1,2.
\end{equation}
Each residual branch is
\begin{equation}
\begin{aligned}
F_\ell(v)
=
\operatorname{Drop}_{0.1}\Bigl(
\operatorname{Linear}_{512\to512}\bigl(
&\operatorname{Drop}_{0.1}\bigl(
\operatorname{GELU}(
\operatorname{Linear}_{512\to512}(v))
\bigr)
\bigr)
\Bigr).
\end{aligned}
\end{equation}
The prediction is
\begin{equation}
\hat{y}(x)
=
\sigma
\left(
\operatorname{Linear}_{512\to1}
\left(
\operatorname{LN}(u^{(2)})
\right)
\right).
\end{equation}

\subsection{Transformer Encoder}

\paragraph{Input tokens.}

The anchor representation is first projected to a 512-dimensional
token:
\begin{equation}
t(x)
=
\operatorname{LN}
\left(
\operatorname{Drop}_{0.1}
\left(
\operatorname{GELU}
\left(
\operatorname{Linear}_{d_0\to512}(\tilde{z}_0(x))
\right)
\right)
\right).
\end{equation}
Let $c_{\mathrm{CLS}}\in\mathbb{R}^{512}$ be a learnable CLS token and
$P\in\mathbb{R}^{2\times512}$ be learnable positional embeddings. The
initial sequence is
\begin{equation}
T^{(0)}
=
\begin{bmatrix}
c_{\mathrm{CLS}}^\top\\
t(x)^\top
\end{bmatrix}
+P.
\end{equation}
The encoder therefore receives one CLS token and one representation
token.

\paragraph{Multi-head self-attention.}

Each layer uses eight attention heads, each with dimension
$512/8=64$. For an input sequence $S$, head $j$ computes
\begin{equation}
\begin{aligned}
Q_j &= \operatorname{Linear}^{Q_j}_{512\to64}(S),\\
K_j &= \operatorname{Linear}^{K_j}_{512\to64}(S),\\
V_j &= \operatorname{Linear}^{V_j}_{512\to64}(S).
\end{aligned}
\end{equation}
Its output is
\begin{equation}
A_j(S)
=
\operatorname{Drop}_{0.1}
\left[
\operatorname{softmax}
\left(
\frac{Q_jK_j^\top}{\sqrt{64}}
\right)
\right]V_j,
\end{equation}
where softmax is applied over the key-token dimension. Head outputs
are concatenated and projected:
\begin{equation}
\operatorname{MHSA}(S)
=
\operatorname{Linear}_{512\to512}
\left(
\operatorname{Concat}
\left(A_1(S),\ldots,A_8(S)\right)
\right).
\end{equation}

\paragraph{Encoder layers.}

Both layers use a pre-norm configuration. For $\ell=1,2$,
\begin{equation}
U^{(\ell)}
=
T^{(\ell-1)}
+
\operatorname{Drop}_{0.1}
\left(
\operatorname{MHSA}_\ell
\left(
\operatorname{LN}^{\mathrm{attn}}_\ell(T^{(\ell-1)})
\right)
\right),
\end{equation}
\begin{equation}
T^{(\ell)}
=
U^{(\ell)}
+
\operatorname{FFN}_\ell
\left(
\operatorname{LN}^{\mathrm{ffn}}_\ell(U^{(\ell)})
\right).
\end{equation}
The feed-forward branch is
\begin{equation}
\begin{aligned}
\operatorname{FFN}_\ell(v)
=
\operatorname{Drop}_{0.1}\Bigl(
\operatorname{Linear}_{2048\to512}\bigl(
&\operatorname{Drop}_{0.1}\bigl(
\operatorname{GELU}(
\operatorname{Linear}_{512\to2048}(v))
\bigr)
\bigr)
\Bigr).
\end{aligned}
\end{equation}

\paragraph{Output head.}

The CLS output of the second encoder layer,
$T^{(2)}_{\mathrm{CLS}}$, is mapped to
\begin{equation}
\hat{y}(x)
=
\sigma
\left(
\operatorname{Linear}_{512\to1}
\left(
\operatorname{LN}(T^{(2)}_{\mathrm{CLS}})
\right)
\right).
\end{equation}

\section{Leave-Anchor-Out Diagnostic}
\label{app:leave-anchor-out}

This experiment fixes CLIP representations as predictor inputs and
compares supervision targets that include or exclude CLIP
neighborhoods. Let $h_0,h_1,h_2,h_3$ denote CLIP, DINOv2, ViT, and
Swin, respectively. The matched control target is
\begin{equation}
y_{\mathrm{anchor}}(x)
=
\frac{1}{3}
\left[
a_k^{(0,1)}(x)
+
a_k^{(0,2)}(x)
+
a_k^{(0,3)}(x)
\right],
\end{equation}
and the leave-anchor-out target is
\begin{equation}
y_{\mathrm{LAO}}(x)
=
\frac{1}{3}
\left[
a_k^{(1,2)}(x)
+
a_k^{(1,3)}(x)
+
a_k^{(2,3)}(x)
\right].
\end{equation}

Both settings use 512-dimensional CLIP inputs, a Residual FFN, the full
training reference set, and $k=50$, with the same training protocol.
They differ only in the supervision target, and a corresponding
predictor is trained for each.

Both targets average three pairwise agreement terms, but
$y_{\mathrm{LAO}}$ contains no CLIP neighborhoods. We compute Pearson
correlation between predictions and the corresponding measured target
to assess predictability when the anchor does not participate in
target construction.

\section{Computational Cost Comparison}
\label{app:cost}

This experiment compares FAISS GPU FlatIP, PyTorch matrix
multiplication followed by Top-$k$ retrieval, and a forward pass through
the Residual FFN. All methods process the same 100,000 ImageNet-1K test
queries individually. Direct measurement uses the full training
reference set and computes neighborhood agreement across CLIP, DINOv2,
and ViT with $k=50$.

Timing starts from precomputed representations. FAISS indexes are
constructed and loaded before the timed loop, and the PyTorch
implementation keeps reference representations on the GPU. After
warm-up, CUDA synchronization is performed before and after processing,
and total elapsed time is recorded.

For $n$ queries processed in $T$ seconds, average per-sample time is
\begin{equation}
t_{\mathrm{sample}}
=
\frac{1000T}{n}
\quad\text{ms/sample}.
\end{equation}
The speedup over direct measurement is
\begin{equation}
\operatorname{Speedup}
=
\frac{T_{\mathrm{measured}}}{T_{\mathrm{predicted}}}.
\end{equation}

Space costs are reported as peak GPU memory and disk storage for
required assets, both in MiB. Direct measurement includes reference
representations or indexes, whereas prediction storage is measured
using the predictor checkpoint.

\section{Association with Downstream Classification Risk}
\label{app:downstream-risk}

\subsection{Linear Classification Probes}

For Aircraft, Caltech101, Cars, DTD, EuroSAT, Food101, Oxford-IIIT Pets,
SUN397, and UCF101, we train separate linear probes on frozen CLIP
representations using the corresponding training splits.

For a dataset with $C$ classes, the probe outputs
\begin{equation}
p(x)
=
\operatorname{softmax}
\left(
W_{\mathrm{cls}}\tilde{z}_0(x)+b_{\mathrm{cls}}
\right),
\end{equation}
where $W_{\mathrm{cls}}\in\mathbb{R}^{C\times d_0}$. Ground-truth
classes are denoted by $c_i$, distinguishing them from agreement
targets $y_i$. The probe training objective is
\begin{equation}
\mathcal{L}_{\mathrm{cls}}
=
-\frac{1}{n_{\mathrm{tr}}}
\sum_{i=1}^{n_{\mathrm{tr}}}
\log p_{c_i}(x_i).
\end{equation}

The disagreement predictor is the CLIP-anchor Residual FFN trained on
ImageNet-1K and remains fixed during downstream evaluation.

\subsection{Multiclass Brier Score}

The sample-wise multiclass Brier score is
\begin{equation}
\label{eq:app-brier}
B_i
=
\sum_{c=1}^{C}
\left(
p_c(x_i)-\mathbf{1}[c=c_i]
\right)^2,
\end{equation}
where $\mathbf{1}[\cdot]$ is the indicator function. Equivalently,
\begin{equation}
B_i
=
\sum_{c=1}^{C}p_c(x_i)^2
-
2p_{c_i}(x_i)
+
1.
\end{equation}
We sum over classes without dividing by the number of classes, giving
$B_i\in[0,2]$. Lower scores indicate that class probabilities are closer
to the one-hot ground-truth label.

\subsection{Kendall Rank Correlation between Disagreement and Brier Risk}

For downstream evaluation only, agreement scores are expressed in the
direction of disagreement:
\begin{equation}
d_i^{\mathrm{measured}}=1-y_i,
\qquad
d_i^{\mathrm{predicted}}=1-\hat{y}_i.
\end{equation}
Predictor training continues to use the original agreement target.

We quantify the association between disagreement and per-sample Brier
risk using Kendall's $\tau_b$, which accounts for tied values. For paired
sequences $d$ and $B$, let $C$ and $D$ denote the numbers of concordant
and discordant sample pairs. Let $T_d$ and $T_B$ denote the numbers of
pairs tied only in $d$ and only in $B$, respectively. Then
\begin{equation}
\tau_b(d,B)
=
\frac{C-D}
{\sqrt{(C+D+T_d)(C+D+T_B)}}.
\end{equation}

For each dataset, the two correlations are
\begin{equation}
\tau_{\mathrm{measured},B}
=
\tau_b(d^{\mathrm{measured}},B),
\qquad
\tau_{\mathrm{predicted},B}
=
\tau_b(d^{\mathrm{predicted}},B).
\end{equation}
A positive value indicates that samples with higher disagreement scores
tend to have higher Brier risk.

Correlations are computed independently within each of the nine
datasets. For $m\in\{\mathrm{measured},\mathrm{predicted}\}$, the
equal-weight average is
\begin{equation}
\bar{\tau}_{m,B}
=
\frac{1}{9}\sum_{j=1}^{9}\tau_{m,B}^{(j)}.
\end{equation}
The unrounded difference on dataset $j$ is
\begin{equation}
\Delta\tau_j
=
\tau_{\mathrm{predicted},B}^{(j)}
-
\tau_{\mathrm{measured},B}^{(j)}.
\end{equation}
Table~\ref{tab:exp6-downstream-risk-pearson} displays correlations to
two decimal places. Its colored differences are calculated from those
displayed values; the Average column is calculated from unrounded
dataset-level correlations before being displayed to two decimal places.

\section{Effect of Reference-Set Size}
\label{app:reference-scaling}

This experiment fixes CLIP as the anchor, DINOv2 and ViT as comparison
models, a Residual FFN predictor, and $k=50$. Predictor training queries
are fixed to 50,000 samples, denoted by $\mathcal{Q}_{\mathrm{tr}}$.

Let $\mathcal{R}_L\subseteq\mathcal{D}_{\mathrm{tr}}$ be a reference
subset containing $L$ samples, where
\begin{equation}
L\in
\{5{,}000,10{,}000,20{,}000,50{,}000,
100{,}000,200{,}000,500{,}000\}.
\end{equation}
Reference subsets are nested and satisfy
\begin{equation}
\mathcal{Q}_{\mathrm{tr}}\cap\mathcal{R}_L
=
\varnothing.
\end{equation}
The full-reference control uses all 1,281,167 training samples.

Let $y^{(L)}(x)$ denote supervision constructed using $\mathcal{R}_L$,
and let $g_{\phi_L}$ denote the corresponding predictor. All predictors
use the same training queries and are selected on validation data and
evaluated on test data against the full-reference target
$y^{(\mathrm{full})}(x)$. Defining
\begin{equation}
\hat{y}_i^{(L)}
=
g_{\phi_L}(\tilde{z}_0(x_i)),
\end{equation}
we compute
\begin{equation}
r_L
=
\operatorname{Corr}_{\mathrm{P}}
\left(
\hat{y}^{(L)},y^{(\mathrm{full})}
\right).
\end{equation}

The full-reference control also uses 50,000 training queries, and its
correlation is denoted by $r_{\mathrm{full}}$. Pearson retention is
\begin{equation}
\operatorname{Retention}(L)
=
\frac{r_L}{r_{\mathrm{full}}}\times100\%.
\end{equation}
When multiple reference subsets are evaluated at the same size, their
correlations are averaged before computing the reported retention.

\end{document}